\documentclass[runningheads]{llncs}
\pdfoutput=1 
 
\usepackage{eccv}

\usepackage{eccvabbrv}

\usepackage{graphicx}
\usepackage{booktabs}

\usepackage[accsupp]{axessibility}  

\usepackage{graphicx}
\usepackage{booktabs}
\usepackage{multirow}
\usepackage{color}

\usepackage[accsupp]{axessibility}  

\usepackage{hyperref}

\usepackage{orcidlink}

\begin{document}

\title{FRI-Net: Floorplan Reconstruction via \\ Room-wise Implicit Representation} 


\author{Honghao Xu\orcidlink{0009-0005-6380-4078} \and
Juzhan Xu\orcidlink{0000-0002-5132-238X} \and
Zeyu Huang\orcidlink{0000-0001-6786-1997} \and \\
Pengfei Xu\orcidlink{0000-0003-4770-4374} \and
Hui Huang\orcidlink{0000-0003-3212-0544} \and
Ruizhen Hu\thanks{Corresponding author}\orcidlink{0000-0002-6798-0336} 
}

\authorrunning{H. Xu et al.}

\institute{Shenzhen University \\
\email{\{littledaisy20001227, juzhan.xu, vcchzy, \\ xupengfei.cg, hhzhiyan, ruizhen.hu\}@gmail.com}
}

\maketitle

\begin{abstract}

In this paper, we introduce a novel method called FRI-Net for 2D floorplan reconstruction from 3D point cloud. 
Existing methods typically rely on corner regression or box regression, which lack consideration for the global shapes of rooms.
To address these issues, we propose a novel approach using a room-wise implicit representation with structural regularization to characterize the shapes of rooms in floorplans.
By incorporating geometric priors of room layouts in floorplans into our training strategy, the generated room polygons are more geometrically regular. We have conducted experiments on two challenging datasets, Structured3D and SceneCAD.
Our method demonstrates improved performance compared to state-of-the-art methods, validating the effectiveness of our proposed representation for floorplan reconstruction.

  \keywords{Floorplan Reconstruction \and Scene Understanding}
\end{abstract}

\section{Introduction}
\label{sec:intro}

Floorplan reconstruction is a fundamental task of scene understanding, and has a variety of applications such as AR/VR, architecture designs, and robotics. Recent works in computer vision tackle the floorplan reconstruction as an image-based detection problem \cite{chen2022heat, chen2019floor, yue2023connecting, su2023slibo}. Given a point cloud of a 3D scene, a top-down image such as density map is extracted as input and the output is a vectorized floorplan, where each room is represented by a polygon, with each vertex of the polygon corresponding to a corner of the room, and the lines connecting the vertices representing the walls of the room.

Previous works on floorplan reconstruction can be divided into corner-based \cite{chen2019floor, yue2023connecting, liu2018floornet, liu2017raster, chen2022heat} methods and box-based \cite{su2023slibo} methods. Corner-based methods represent the floorplan as a collection of corner points and the edges between them. These methods achieve floorplan reconstruction by detecting corners and predicting the connections between them. However, when the input signals are noisy or missing, predictions in the affected areas tend to be inaccurate, which impacts the shape of the reconstructed rooms. This limitation demonstrates a disregard for the global room shape and a lack of robustness. Recently, SLIBO-Net \cite{su2023slibo} introduced a box-based representation for floorplans, where the floorplan is sliced into multiple boxes. This representation allows for a more direct consideration of the local shape information of each room through box regression; however, it cannot represent non-Manhattan shapes.

\begin{figure*}[!t]
    \centering
    \includegraphics[width= 1.0\textwidth]{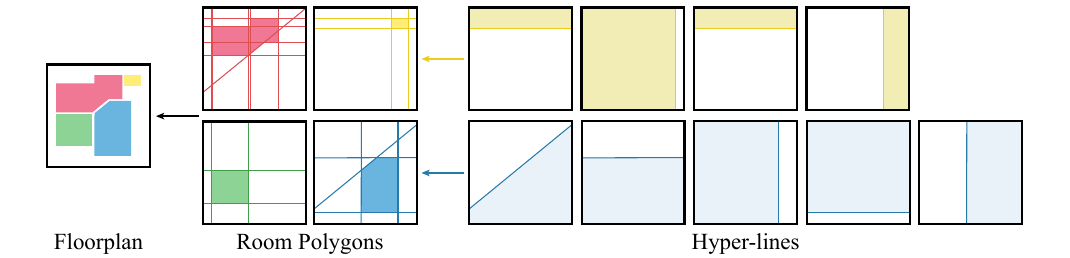}
    \caption{Floorplan representation. A floorplan can be represented as an assembly of room polygons, with each polygon modeled through a room-wise implicit representation. The room-wise implicit representation can be constructed using a set of hyper-lines.}
    \label{fig:floorplan_representation}
\end{figure*}

In this work, we propose a novel approach for \textbf{F}loorplan reconstruction via \textbf{R}oom-wise \textbf{I}mplicit representation, coined FRI-Net. As shown in Figure \ref{fig:floorplan_representation}, a floorplan can be represented as an assembly of room polygons. Our key idea is to represent each room as an implicit function: given \(n\) point coordinates as input, this function outputs \(n\) values indicating whether these points are inside or outside the corresponding room. Our key observation is that the room polygons are composed of a set of lines that intersect to form the rooms' geometries. Inspired by BSP-Net \cite{chen2020bsp}, we develop a learning framework that utilizes these explicit line parameters to construct the room-wise implicit representation. Based on the geometric priors of the floorplan rooms, where the external wall predominantly consists of axis-aligned lines, we propose a two-phase training strategy as the structural regularization to facilitate the learning of room geometries: first focusing on horizontal and vertical lines to establish a robust initial structure, then introducing diagonal lines for angular features. Compared to previous methods, our room-wise structural implicit representation allows for a more direct consideration of the global shape feature of the rooms, enabling adaptive completion of noisy or missing areas and ensuring shape consistency.


To validate the effectiveness of our proposed method, we evaluate our methods on two challenging datasets, Structured3D \cite{zheng2020structured3d} and SceneCAD \cite{avetisyan2020scenecad}. The experimental results indicate that our method yields a significant improvement in the quality of the reconstructed floorplan compared with the state-of-the-art. 

To summarize, our contributions are as follows:
\begin{itemize}
\item[$\bullet$] We propose to use a room-wise implicit representation with structural regularization to characterize the global shapes of rooms in a floorplan.
\item[$\bullet$] We have developed a framework, named FRI-Net, that implements this proposed representation for floorplan reconstruction.
\item[$\bullet$] We validate that our method can achieve high-quality reconstruction results compared with the state-of-the-art.
\end{itemize}

\section{Related work}

\subsection{Classical methods}

Floorplan reconstruction has been widely studied in computer vision. Early methods relied on low-level image processing techniques such as hough transforms or histograms for floorplan reconstruction \cite{adan20113d, okorn2010toward, llados1997system, budroni2010automated, monszpart2015rapter, sanchez2012planar}. For example, Okorn et al. \cite{okorn2010toward} projected the points from walls to obtain a 2D density histogram, from which line segments were extracted using the hough transform to construct the floorplan structure. Similarly, Adan et al. \cite{adan20113d} applied the hough transform to detect and identify the position of walls. Beyond these foundational image processing techniques, some methods have begun to employ more sophisticated technologies, such as graph models \cite{cabral2014piecewise, furukawa2009reconstructing, ikehata2015structured}. For instance, Ikehata et al. \cite{ikehata2015structured} characterized the entire floorplan in the form of a graph, where each node corresponds to structural elements such as rooms and walls. Cabral et al. \cite{cabral2014piecewise} proposed to transform the floorplan reconstruction problem into solving the shortest path problem on a specially designed graph. However, these methods are hard to generalize to new scenes since they rely heavily on heuristics.

\subsection{Deep learning-based methods}
With the development of deep learning, more recent works have begun to leverage neural networks for floorplan reconstruction. Bottom-up processing is a popular approach for floorplan reconstruction \cite{liu2017raster, liu2018floornet, chen2022heat}. Floor-Net \cite{liu2018floornet} first detected room corners, then used Integer Programming \cite{schrijver1998theory} to assemble them to generate the final vectorized floorplan. However, they need to restrict the solution space of the Integer Programming to Manhattan scenes. Recently, HEAT \cite{chen2022heat} proposed an attention-based neural network for structural reconstructions, where they first detected corners and classified edge candidates between corners in an end-to-end manner. However, these bottom-up processes all suffer from missing corners, they can not recover edges from undetected corners which results in missing walls. By taking the geometric constraint of the floorplan into consideration, where the external walls of a room must form a closed 1D loop, recent methods begin to formulate the floorplan reconstruction as the generation of multiple polygon loops, one for each room \cite{chen2019floor, stekovic2021montefloor, yue2023connecting}. Floor-SP \cite{chen2019floor} started from room segments generated by Mask-RCNN \cite{he2017mask}, then they sequentially reconstructed the polygon of each segment by solving a shortest path problem.
MonteFloor \cite{stekovic2021montefloor} took a similar framework which started from room segmentation, then they adapted Monte Carlo Tree Search \cite{browne2012survey} to assemble room proposals into the final graph of floorplan. The reconstruction results of the above two-stage pipelines rely on the quality of room segmentation and the effectiveness of post-optimization techniques. Yue et al. \cite{yue2023connecting} introduced an end-to-end transformer architecture that directly predicts a set of polygons in a single stage. This approach generates each room as a variable-length sequence of ordered vertices which leverages the sequence prediction capabilities of Transformer \cite{vaswani2017attention}. However, such floorplan representation fails to consider the shape of the room, which will cause the generation of irregular room polygons. Recently, Chen et al. \cite{chen2024polydiffuse} proposed to formulate floorplan reconstruction as a generation process and introduced a Guided Set Diffusion Model \cite{ho2020denoising} to optimize the room representation. 
The above corner-based methods achieve floorplan reconstruction through discrete corner regression, which ignores the global shape information of the rooms. To address this limitation, Su et al. \cite{su2023slibo} employed a box-based representation that slices the floorplan into a set of boxes. Such a method achieves floorplan reconstruction through box regression and allows for a consideration of local room shape. Compared with the above methods, we consider the global shape of rooms by directly learning their implicit function.

\section{Method}
\begin{figure*}[!t]
    \centering
    \includegraphics[width= 1.0\textwidth]{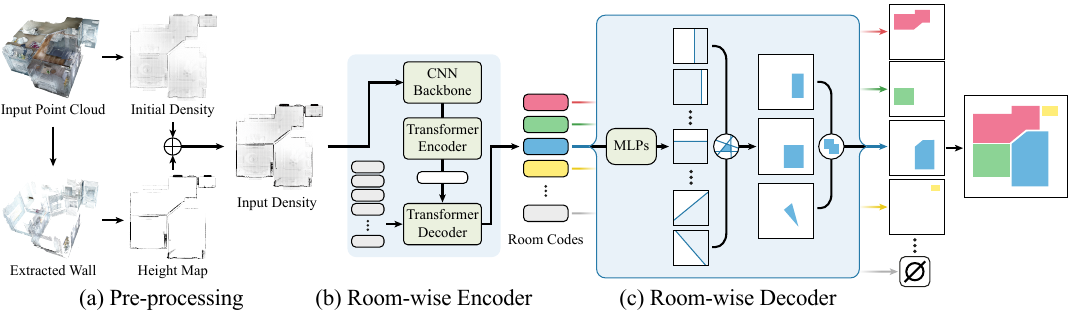}
    \caption{Overview. (a) Given the input 3D point cloud, we use the pre-processing module to obtain the input image by considering the density of the floorplan and the height of the extracted wall. (b) For the input image, we use the room-wise encoder to output several room feature codes, each corresponding to a latent room representation; 
    (c) For each room code, we use the room-wise decoder to output the corresponding room polygon. Multiple polygons are merged to form the final vectorized floorplan.}
    \label{fig:overview}
\end{figure*}

Figure \ref{fig:overview} illustrates the overview of FRI-Net, which is divided into three main parts: (a) pre-processing (Section \ref{sec:pre-processing}): given the input 3D point cloud, we extract its wall area to obtain the wall's height map, which is then integrated with the density map to obtain the input image; (b) room-wise encoder (Section \ref{sec:room-wise encoder}): for the input image, we use a room-wise encoder to output several room feature codes, each corresponding to a latent representation of a room; (c) room-wise decoder (Section \ref{sec:room-wise decoder}): for each room feature code, we use a room-wise decoder to output the corresponding room polygon. Multiple polygons are merged to form the final vectorized floorplan.

\subsection{Pre-processing}
\label{sec:pre-processing}

To estimate the wall structure of a floorplan from the input point cloud, existing methods typically project the point cloud along the gravity axis to obtain the density map. This process is employed because the number of points in the wall area of the point cloud tends to be higher, making the walls appear clearer in the density map. However, a limitation of this method is the loss of height information of the point cloud during the projection process. Our observation is that some walls, despite having weaker density signals in certain areas after projection due to missing parts of the point cloud, still retain clear height information. Therefore, in addition to considering the density information of the input point cloud, we must also consider the height information of the walls. 

Specifically, we first estimate the normals of the input point cloud. Based on the normals' direction, we apply the region growing algorithm \cite{marshall2001robust} to segment the point cloud into multiple areas. This is beneficial for isolating walls, as wall regions are relatively continuous in space and their normals are relatively consistent. Subsequently, we use the RANSAC algorithm \cite{fischler1981random} to extract planes from each segmented area. Based on the plane parameters (perpendicular to the xy plane) of each area, we extract the point cloud of the wall parts, as demonstrated in Figure \ref{fig:overview}(a). Finally, by calculating the maximum height of the wall areas, we obtain a height map and combine it with the density map to obtain the final input image.

\subsection{Room-wise Encoder}
\label{sec:room-wise encoder}

Since our method views floorplan reconstruction as the reconstruction of multiple rooms, our initial step is to obtain a feature representation for each room. Following \cite{yue2023connecting, su2023slibo, chen2024polydiffuse}, we employ the DETR-based \cite{carion2020end, zhu2020deformable} transformer architecture to obtain feature representations for each room. 
As shown in Figure \ref{fig:overview}(b), we first input images into a CNN backbone to extract image feature maps. 
These feature maps are subsequently processed by a transformer encoder, yielding enhanced image features.
Then, we input $m$ learnable embeddings into the transformer decoder. These embeddings adaptively extract local room features from the global image features output by the transformer encoder, such that each output embedding corresponds to a latent representation of a room. These output embeddings 
are then independently fed into the room-wise decoder for further processing, ultimately generating the corresponding room polygon.

\subsection{Room-wise Decoder}
\label{sec:room-wise decoder}

\begin{figure*}[!t]
    \centering
    \includegraphics[width= 1.0\textwidth]{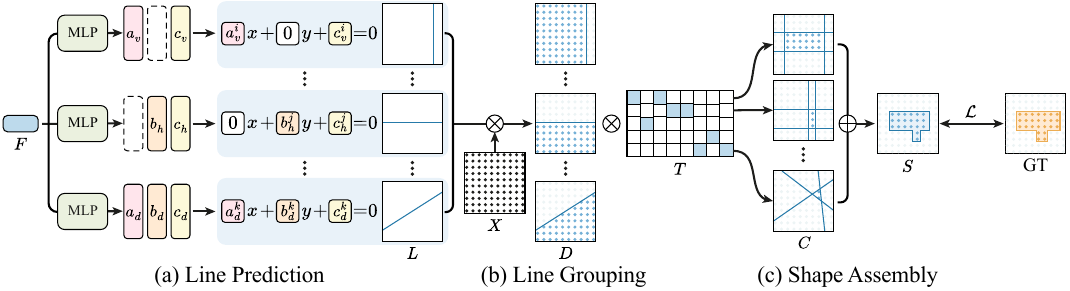}
    \caption{Room-wise decoder. The room-wise decoder takes the room feature code \(F\) and \(n\) query points \(X\) as inputs and outputs the occupancy \(S\) of these points, indicating whether the points are inside or outside the corresponding room.}
    \label{fig:room-wise decoder}
\end{figure*}

As outlined in Section \ref{sec:intro}, we propose employing a room-wise implicit representation to characterize the shape of rooms. We achieve our goal via a room-wise decoder. As illustrated in Figure \ref{fig:room-wise decoder}, for each room code \(F \in \mathbb{R}^{q} \), we input it into the room-wise decoder along with \(n\) query points \(X \in \mathbb{R}^{n \times 3}\), where each row represents the coordinates of one point \((x, y, 1)\). The room-wise decoder then outputs the occupancy \(S \in \mathbb{R}^{n \times 1}\) of these points, indicating whether the points are inside or outside the corresponding room. 
To integrate the explicit lines into the implicit representation, inspired by BSP-Net \cite{chen2020bsp}, the room-wise decoder comprises three core modules: line prediction, line grouping, and shape assembly. In the line prediction module, we predict the lines that constitute the room's polygon. In the line grouping module, the predicted lines are grouped together, with each group forming a convex primitive. In the shape assembly module, all the convex primitives are merged to form the shape of the room polygon.

\subsubsection{Line Prediction.}

A 2D line can be represented by parameters \(a\), \(b\), and \(c\) using the equation \(ax + by + c = 0\). Considering that the wall boundary of a room in a floorplan is always axis-aligned, we focus on directly predicting three types of lines: horizontal, vertical, and diagonal. Given the room's feature code \(F \in \mathbb{R}^{q}\), three distinct MLPs are employed to generate parameters for these specific line types. For vertical lines, one MLP produces parameters \(L_{\text{vertical}} = [a_{v}, 0, c_{v}] \in \mathbb{R}^{l\times3}\), where \(b_{v} = 0\) to ensure vertical alignment. For horizontal lines, another MLP outputs \(L_{\text{horizontal}} = [0, b_{h}, c_{h}] \in \mathbb{R}^{l\times3}\), with \(a_{h} = 0\) to maintain horizontal orientation. 
Lastly, for diagonal lines, an MLP generates \(L_{\text{diagonal}} = [a_{d}, b_{d}, c_{d}] \in \mathbb{R}^{l\times3}\), allowing for the representation of lines at various angles. 
To streamline notation, we aggregate these line parameters into a single matrix \(L = [L_{\text{horizontal}}, L_{\text{vertical}}, L_{\text{diagonal}}] \in \mathbb{R}^{3l\times3}\), representing the combined set of lines. 

\subsubsection{Line Grouping.}
\label{sec:line grouping}
To facilitate the learning of complex room geometries, we first combine the predicted lines into convex primitives. We employ a binary selection matrix \(T \in \mathbb{R}^{3l \times u}\) to split the predicted lines into $u$ distinct groups, 
where \( T(i,j) = 1 \) indicates that the \( i\)-th line is assigned to the \( j \)-th group.
Subsequently, for each group, we apply intersection operation on the inner regions partitioned by these lines to construct convex primitives. For \(n\) query points \(X \in \mathbb{R}^{n \times 3}\), 
we first compute the approximate signed distances from these points to the lines: \(D = X L^T \in \mathbb{R}^{n \times 3l}\).
We then determine the inside-outside status of all points for each convex primitive using the following equation:
\begin{equation}
    C = \text{ReLU}(D) T \quad\left\{\begin{array}{ll}
=0 & \text { in } \\
>0 & \text { out }
\end{array}\right. .
\label{eq:convex}
\end{equation}
The \(i\)-th point is considered inside the \(j\)-th primitive if and only if \(C(i, j) = 0\). This multiplication operation effectively represents the intersection of the inner regions partitioned by the above predicted lines.

\subsubsection{Shape Assembly.}

After obtaining all the convex primitives, we combine these primitives to form the shape of the room. We can determine the occupancy of the $i$-th point with respect to the room polygon by applying a min-pooling operation on the $i$-th row of \(C \in \mathbb{R}^{n \times u}\):
\begin{equation}
{{S^{*}}}(i)=\min_{1\le j \le u}\left(C(i,j)\right) \quad\left\{\begin{array}{ll}
=0 & \text { in } \\
>0 & \text { out }
\end{array}\right.
\label{eq:s_accurate}    
\end{equation}
The expression above represents the union operation across all convex primitives: a point is considered to be inside the room if it is inside at least one of the primitives. Following BSP-Net \cite{chen2020bsp}, to facilitate training, we adopt a weighted sum approach among all primitives. This approach allows gradients to be backpropagated among them, with the expression as follows:
\begin{equation}
{S^{+}}(i)=\left[\sum_{1\le j \le u} W(j)\left[1-C(i,j)\right]_{[0,1]}\right]_{[0,1]}\left\{\begin{array}{l}
=1 \approx \text { in } \\
{< 1 \approx \text { out }}
\end{array}\right.
\label{eq:s_approximate}    
\end{equation}
where \(W \in \mathbb{R}^{u \times 1}\) is a weight vector and \({\left[\cdot \right]}_{\left[0, 1 \right]}\) denote clipping values into $[0, 1]$. It is worth noting that the expression here is just an approximation, since small values corresponding to points outside could accumulate to a sum exceeding 1, misclassifying an external point as inside.

\subsection{Loss function}

\subsubsection{Bipartite matching.}

Our room-wise decoder generates \(m\) sets of occupancy values, \(S \in \mathbb{R}^{m \times n \times 1}\), which indicate whether the query points \(X \in \mathbb{R}^{n \times 3}\) are inside or outside these \(m\) predicted rooms. However, the number of ground truth (GT) room polygons, denoted as \(m_{gt}\), varies for each floorplan. This variability requires us to establish a matching relationship between the predicted occupancy \(S\) and the GT occupancy \(S_{gt} \in \mathbb{R}^{m_{gt} \times n \times 1}\) to facilitate training. 

Following \cite{carion2020end, yue2023connecting}, we address the correspondence between predicted and GT occupancy by solving a bipartite matching problem. Specifically, we pad the GT occupancy \(S_{gt}\) to match the dimensions of the predicted occupancy, resulting in a padded GT 
\(\hat{S}_{gt} \in \mathbb{R}^{m \times n \times 1}\). In this padding process, we fill the added elements with zeros, representing invalid rooms (\(\forall i \in [m_{gt}+1, m], \hat{S}_{gt}^i=\textbf{0}\in \mathbb{R}^{n \times 1}\)). We then find a bipartite matching between the predicted occupancy and the padded GT occupancy by searching for a permutation $\hat{\sigma}$ with minimal cost:
\begin{equation}
\hat{\sigma}=\underset{\sigma}{\arg \min } \sum_{i=1}^{m} \mathcal{D}\left(S(i), \hat{S}_{gt}({\sigma(i)})\right)
\label{eq:matching}
\end{equation}
where \(\mathcal{D}(\cdot)\) represents the discrepancy between the predicted occupancy $S(i)$ and the GT occupancy $\hat{S}_{gt}(\sigma(i))$, which we refer to as the reconstruction loss. Finally, we solve for the permutation \({\sigma}^*\) that minimizes the cost using the Hungarian algorithm, thereby establishing the correspondence between the predicted occupancy and the GT occupancy. The measure of the reconstruction loss varies depending on the training phase, which will be discussed in the following section.

\subsubsection{Loss function.}

As mentioned in Section \ref{sec:room-wise decoder}, given that \(S^+\) offers better gradient backpropagation compared to \(S^*\), it allows us to optimize the shapes of all convex primitives simultaneously, however, \(S^+\) is only approximate. Furthermore, since we need to learn a discrete binary matrix representing the combination relationships between predicted lines, directly optimizing this matrix is quite challenging. Therefore, considering all these factors, we need to design multiple loss functions for various training phases to achieve finer reconstruction results.

In the early stages of training, to facilitate better gradient backpropagation, we utilize \(S^+\) to represent the predicted occupancy, the loss function is:
\begin{equation}
    \mathcal{L}^+ = \mathcal{L}_{rec}^+ + \mathcal{L}_{T}^+ + \mathcal{L}_{W}^+.
    \label{eq:L_stage0}
\end{equation}
\(\mathcal{L}_{rec}^+\) represents the reconstruction loss, we calculate the least squares loss between \(S^+\) and the padded GT occupancy \(\hat{S}_{gt}\) based on the permutation \(\sigma^*\):
\begin{equation}
    \mathcal{L}_{rec}^+ = \frac{1}{m} \sum_{i=1}^{m} \mathbb{E} \left[\left( S^+(i) - \hat{S}_{gt}({{\sigma^*(i)})}\right)^2\right].
\end{equation}
We keep the weights of the binary matrix \(T\) as continuous and enforce its elements to be bounded within the range \([0,1]\):
\begin{equation}
    \mathcal{L}_{T}^{+}=\sum_{t \in {T}} \max (-t, 0)+\sum_{t \in {T}} \max (t-1,0).
\end{equation}
To prevent gradient vanishing caused by clipping, each element of \(W\) is initialized to 0 and gradually approximates to 1 through the following loss:
\begin{equation}
    \mathcal{L}_{W}^+ = \sum_{i} |W(i) - 1|.
\end{equation}

In the final stage of training, to achieve a finer reconstruction of rooms, we use \(S^*\) to represent the inside-outside status of query points. Additionally, we impose extra constraints on the selection matrix to encourage its discretization, the loss function is:
\begin{equation}
    \mathcal{L}^* = \mathcal{L}_{rec}^* + \mathcal{L}_{T}^*,
    \label{eq:L_stage1}
\end{equation}
where \(\mathcal{L}_{rec}^*\) is the reconstruction loss as below:
\begin{equation}
    \begin{aligned}
        \mathcal{L}_{rec}^* = \frac{1}{m} \sum_{i=1}^{m} \mathbb{E} \left[\hat{S}_{gt}({{\sigma^*(i)})} \cdot \max(S^*(i), 0)\right] \\ 
        + \frac{1}{m} \sum_{i=1}^{m} \mathbb{E} \left[(1 - \hat{S}_{gt}({{\sigma^*(i)})}) \cdot (1 - \min(S^*(i), 1))\right]
    \end{aligned}
\end{equation}
The above equation encourages points outside the room to be one and points inside the room to be zero. To further discretize \(T\), we choose a threshold \(\gamma=0.01\) and pull elements smaller than \(\gamma\) towards 0, and encourage elements larger than \(\gamma\) to be 1 through the following loss:
\begin{equation}
    \mathcal{L}_T^* = \sum_{t\in T}\mathbf{1}\left(t < \gamma \right)|t| + \sum_{t\in T}\mathbf{1}\left( t \ge \gamma\right)|t-1|.
\end{equation}
The training details will be fully discussed in Section \ref{sec: expermient setting}.

During inference, we employ \(\gamma=0.01\) to discretize \(T\) into binary values (setting \(t = (t > \gamma) ? 1 : 0\)), and then we use the predicted lines \(L\) along with the selection matrix \(T\) to construct the output polygon. Since we penalize invalid room feature codes during training with all-zero occupancy values, during inference, we can directly determine the validity of each room polygon based on the output occupancy values.

\section{Experiments}

\subsection{Experiments Setting}
\label{sec: expermient setting}
\subsubsection{Implementation details.}
For the room-wise encoder, consistent with previous work \cite{chen2022heat, yue2023connecting, su2023slibo}, we utilize ResNet-50 \cite{he2016deep} as the CNN backbone. The entire transformer comprises 6 encoder layers and 7 decoder layers, each with 256 channels. The number of room feature codes $m$ is set to 20. We sample 4096 points uniformly in the input image as the input query points during training. The number of horizontal/vertical/diagonal lines $l$ is set to 256. The number of convex primitives $u$ is set to 64.

\subsubsection{Training details.}
Based on the geometric priors of floorplan rooms, where the external walls predominantly consist of horizontal and vertical lines, we propose a training strategy with three stages. Initially, we focus on optimizing horizontal and vertical lines using Loss \ref{eq:L_stage0}, conducting this phase over 600 epochs with a batch size of 16. In this stage, diagonal lines are not involved in the construction of the room shapes. This step aims to capture the basic shape of the rooms leveraging axis-aligned lines. Subsequently, upon establishing the foundational room shapes, we incorporate diagonal lines into the construction of room shapes. This phase involves joint training of all lines—horizontal, vertical, and diagonal—again using Loss \ref{eq:L_stage0}, for an additional 600 epochs at the same batch size of 16. The inclusion of diagonal lines is intended to refine the representation of room shapes, accommodating for the angles area. In the final phase, we proceed to optimize all lines with Loss \ref{eq:L_stage1} for 600 epochs with a batch size of 16. This comprehensive approach ensures a gradual refinement of the room shapes, starting from the most significant geometric features to the finer details, resulting in a detailed and accurate reconstruction of the floorplan rooms.

We implement our model in PyTorch and train it on an NVIDIA GeForce RTX 3090 GPU. We use the Adam optimizer \cite{kingma2014adam} with a weight decay factor of 1e-4, setting the learning rate to 2e-4 and decaying it by a factor of 0.1 in the last 30\% of the epochs.
\subsubsection{Dataset and evaluation metrics.}
We conduct experiments on two large indoor synthetic datasets: Structured3D \cite{zheng2020structured3d} and SceneCAD \cite{avetisyan2020scenecad}. For the Structured3D dataset, consistent with previous work \cite{yue2023connecting, chen2022heat}, we divide the entire dataset into 3,000 training samples, 250 validation samples, and 250 test samples. For the SceneCAD dataset, annotations are only available for the training and validation splits. Therefore, following RoomFormer \cite{yue2023connecting}, we train our model on its training set and report experimental metrics on the validation set. We process the raw point cloud through the pre-processing module mentioned in Section \ref{sec:pre-processing}, resulting in images of a \(256\times 256\) dimension. The pixel values of the images are normalized to the range \([0, 1]\). Consistent with previous methods \cite{stekovic2021montefloor, chen2022heat, yue2023connecting, su2023slibo, chen2024polydiffuse}, we report Precision, Recall, and F1 scores across three geometric aspects: Room, Corner, and Angle.

\subsubsection{Baselines.}
We compare our method against seven approaches: LETR \cite{xu2021line}, HEAT \cite{chen2022heat}, Floor-SP \cite{chen2019floor}, MonteFloor \cite{stekovic2021montefloor}, SLIBO-Net \cite{su2023slibo}, RoomFormer \cite{yue2023connecting}, and PolyDiffuse \cite{chen2024polydiffuse}. LETR \cite{xu2021line} is a line segment detection framework, which is adapted for floorplan reconstruction in \cite{chen2022heat}. HEAT \cite{chen2022heat} first detects corners and then determines the connectivity between corners based on the neural network. Both Floor-SP \cite{chen2019floor} and MonteFloor \cite{stekovic2021montefloor} initially use Mask-RCNN \cite{he2017mask} for room segmentation, followed by learning-free optimization techniques for room vectorization. SLIBO-Net \cite{su2023slibo} represents floorplans as multiple boxes, with post-processing employed to recover floorplans from these boxes. RoomFormer \cite{yue2023connecting} leverages the transformer architecture to simultaneously output multiple corner sequences for floorplan reconstruction. PolyDiffuse \cite{chen2024polydiffuse} views floorplan reconstruction as a generation process conditioned on sensor data, utilizing a diffusion model \cite{ho2020denoising}. This approach requires a proposal generator to provide initial reconstruction results for further processing.

\subsection{Comparisons to state-of-the-art methods}
\begin{table}[!t]
\centering
\caption{Quantitative comparison on Structured3D \cite{zheng2020structured3d} test set. Results of prior work are copied from \cite{yue2023connecting, su2023slibo, chen2024polydiffuse}. Runtime is averaged over the test set. We rename PolyDiffuse \cite{chen2024polydiffuse} as PD for simplification.}

\setlength{\tabcolsep}{3pt}
\begin{tabular}{c c  ccc  ccc ccc}
\hline
\multirow{2}{*}{Methods} 
& 
& \multicolumn{3}{c}{Room} 
& \multicolumn{3}{c}{Corner} 
& \multicolumn{3}{c}{Angle} \\ 
\cmidrule(r){3-5} \cmidrule(r){6-8} \cmidrule(r){9-11}
       &t (s) & Prec.   & Rec.   & F1    & Prec.    & Rec.   & F1     & Prec.   & Rec.   & F1     \\ \hline
LETR \cite{xu2021line}    & 0.04     & 94.5    & 90.0   & 92.2  & 79.7     & 78.2   & 78.9   & 72.5    & 71.3   & 71.9   \\
Floor-SP \cite{chen2019floor}   &785   & 89.0    & 88.0   & 88.0  & 81.0     & 73.0   & 76.0   & 80.0    & 72.0   & 75.0   \\
MonteFloor \cite{stekovic2021montefloor} &71    & 95.6    & 94.4   & 95.0  & 88.5     & 77.2   & 82.5   & 86.3    & 75.4   & 80.5   \\
HEAT \cite{chen2022heat}  &0.11       & 96.9    & 94.0   & 95.4  & 81.7     & 83.2   & 82.5   & 77.6    & 79.0   & 78.3   \\
SLIBO-Net \cite{su2023slibo}  &0.17  & 99.1    & 97.8   & 98.4  & 88.9     & 82.1   & 85.4   & 87.8    & 81.2   & 84.4   \\ 
RoomFormer \cite{yue2023connecting} &\textbf{0.01}   & 97.9    & 96.7   & 97.3  & 89.1     & \textbf{85.3}   & 87.2   & 83.0    & 79.5   & 81.2   \\
FRI-Net   &0.09       & \textbf{99.5}    & \textbf{98.7}   & \textbf{99.1}  & \textbf{90.8}     & 84.9   & \textbf{87.8}   & \textbf{89.6}    & \textbf{84.3}   & \textbf{86.9}   \\ \hline
RoomFormer+PD \cite{chen2024polydiffuse} &- & 98.7    & 98.1   & 98.4  & 92.8     & \textbf{89.3}   & 91.0   & 90.8    & \textbf{87.4}   & 89.1   \\ 
FRI-Net+PD \cite{chen2024polydiffuse} &- & \textbf{99.6}    & \textbf{98.6}   & \textbf{99.1}  & \textbf{94.2}     & 88.2   & \textbf{91.1}   & \textbf{91.9}    & 86.7   & \textbf{89.2}   \\ \hline
\end{tabular}
\label{tab:stru3d}
\end{table}
\begin{table}[!t]
\centering
\caption{
Comparison on SceneCAD \cite{avetisyan2020scenecad}.
Results of prior work are copied from \cite{yue2023connecting}.}
\setlength{\tabcolsep}{4pt}
\begin{tabular}{c c c ccc ccc}
\hline
\multirow{2}{*}{Methods} &
           & Room & \multicolumn{3}{c}{Corner} & \multicolumn{3}{c}{Angle} \\ 
            \cmidrule(r){3-3} \cmidrule(r){4-6} \cmidrule(r){7-9}
    &t (s) & IoU  & Prec.    & Rec.   & F1     & Prec.   & Rec.   & F1     \\ \hline
Floor-SP \cite{chen2019floor} &26   & 91.6 & 89.4     & 85.8   & 87.6   & 74.3    & 71.9   & 73.1   \\
HEAT \cite{chen2022heat}&0.12      & 84.9 & 87.8     & 79.1   & 82.5   & 73.2    & 67.8   & 70.4   \\
RoomFormer \cite{yue2023connecting} & \textbf{0.01} & 91.7 & 92.5     & 85.3   & 88.8   & 78.0    & \textbf{73.7}   & 75.8   \\
FRI-Net     &0.07  & \textbf{92.3} & \textbf{92.8}     & \textbf{85.9}   & \textbf{89.2}   & \textbf{78.3}    & 73.6   & \textbf{75.9}   \\ \hline
\end{tabular}
\label{tab:scenecad}
\end{table}

\subsubsection{Quantitative evaluation.}

Quantitative comparisons to state-of-the-art floorplan reconstructions on Structured3D \cite{zheng2020structured3d} are shown in Table.\ref{tab:stru3d}. After incorporating PolyDiffuse \cite{chen2024polydiffuse} for post-optimization, the performance of both our method and RoomFormer \cite{yue2023connecting} has improved. Without such a post-process, our approach achieves the best results across three geometric levels, with an improvement of +1.8 in Room F1 score, +0.6 in Corner F1 score, and notably, +5.7 in Angle F1 score. Given our reconstruction quality outperforms RoomFormer \cite{yue2023connecting}, our method continues to achieve superior performance across all three levels with PolyDiffuse \cite{chen2024polydiffuse} incorporated for post-optimization. Furthermore, despite our method not explicitly predicting corners during the optimization process, it still maintains a comparable performance in corner recall.
For the runtime comparison, our method is slightly slower than RoomFormer\cite{yue2023connecting} since we require post-processing to filter out the lines that constitute the geometry of rooms. We believe that our method achieves a good balance between performance and computational cost.

On the SceneCAD \cite{avetisyan2020scenecad} dataset, we compare our method with Floor-SP \cite{chen2019floor}, HEAT \cite{chen2022heat}, and RoomFormer \cite{yue2023connecting} (for which the source code is publicly available). Given that this dataset essentially contains only one room, we evaluate the reconstruction quality of room shape
using IoU. As shown in Table.\ref{tab:scenecad}, our method still achieves the best performance across nearly all evaluation metrics.

\subsubsection{Qualitative evaluation.}

\begin{figure*}[!t]
    \centering
    \includegraphics[width= 1.0\textwidth]{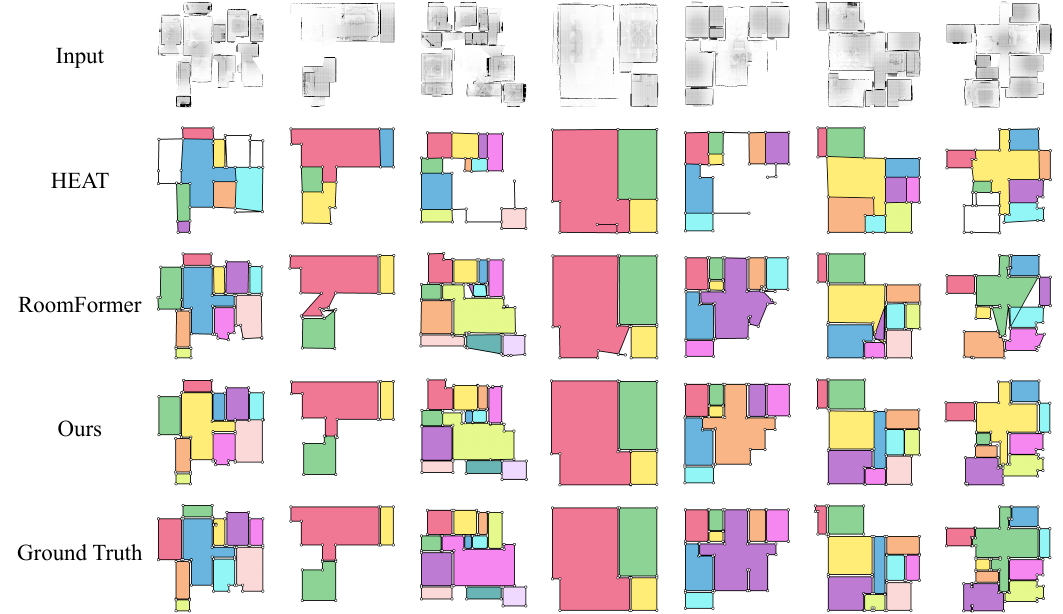}
    \caption{Qualitative evaluations on Structured3D \cite{zheng2020structured3d}.}
    \label{fig:exp_stru3d}
\end{figure*}

\begin{figure*}[!t]
    \centering
    \includegraphics[width= 1.0\textwidth]{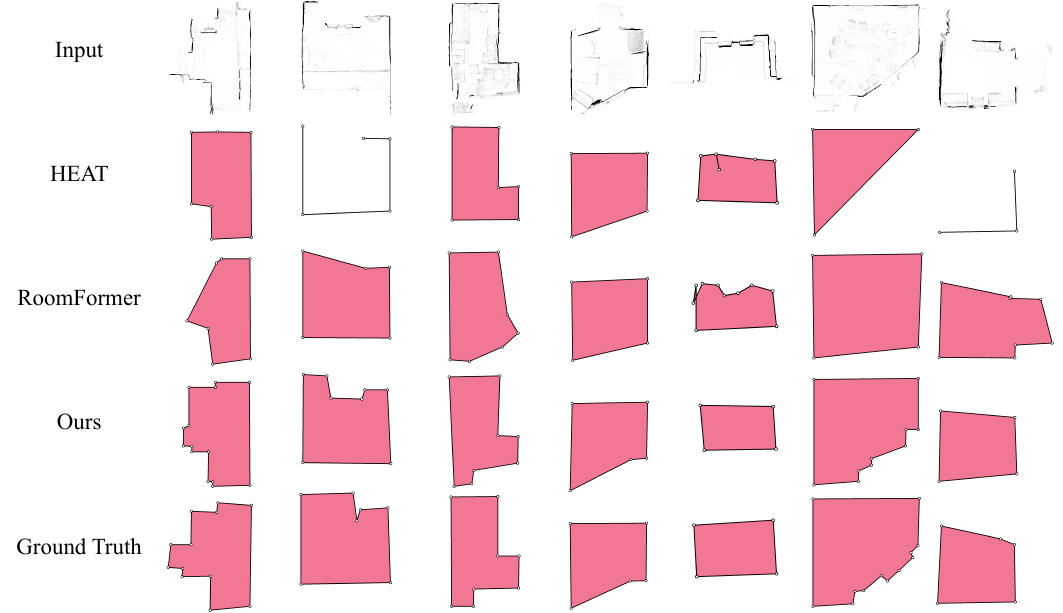}
    \caption{Qualitative evaluations on SceneCAD \cite{avetisyan2020scenecad}.}
    \label{fig:exp_scenecad}
\end{figure*}

Visual comparisons on Structured3D \cite{zheng2020structured3d} are presented in Figures \ref{fig:exp_stru3d}. HEAT \cite{chen2022heat} adopts a two-stage pipeline that first detects corners and then judges their connectivity. It encounters several issues including missing corners (\(3^{\text{rd}}, 5^{\text{th}}\) row), redundant corners (\(1^{\text{st}}, 4^{\text{th}}, 7^{\text{th}}\) row), and incorrect edge connectivity (\(2^{\text{nd}}\) row). 
RoomFormer \cite{yue2023connecting} approaches floorplan reconstruction by predicting sequences of polygon corners. By sequentially connecting corners, it ensures the generation of a closed polygon. However, due to the lack of effective constraints on the relationships between corners within the sequence, the shapes of the polygons may be irregular, as shown in the \(3^{\text{rd}}\) row of Figure \ref{fig:exp_stru3d}.
Our method, by explicitly incorporating the geometric priors of floorplans into the construction of floorplan representations, results in more regular room shapes. Moreover, by employing a neural implicit representation to characterize the shape of rooms, where the boundaries of rooms are expressed as the zero-level-set of an occupancy field, we ensure that our generated polygons are closed. Figure \ref{fig:exp_scenecad} further demonstrates the same pattern, with our method producing more plausible results compared to corner-based optimization methods. 

We have extended our method toward semantically-rich floorplans, please see the supplementary for more details.

\begin{table}[!t]
\centering
\caption{
Cross-data generalization.}
\setlength{\tabcolsep}{4pt}
\begin{tabular}{c c ccc ccc}
\hline
\multirow{2}{*}{Methods}
           & Room & \multicolumn{3}{c}{Corner} & \multicolumn{3}{c}{Angle} \\ 
            \cmidrule(r){2-2} \cmidrule(r){3-5} \cmidrule(r){6-8}
    & IoU  & Prec.    & Rec.   & F1     & Prec.   & Rec.   & F1     \\ \hline
HEAT \cite{chen2022heat}      & 52.5 & 50.9     & 51.1   & 51.0   & 42.2    & 42.0   & 41.6   \\
Roomformer \cite{yue2023connecting} & 74.0 & 56.2     & 65.0   & 60.3   & 44.2    & 48.4   & 46.2   \\
FRI-Net       & \textbf{80.6} & \textbf{66.4}     & \textbf{79.5}   & \textbf{72.4}   & \textbf{56.3}    & \textbf{67.2}   & \textbf{61.3}   \\ \hline
\end{tabular}
\label{tab:generalize}
\end{table}
\subsubsection{Cross-data generalization.}
We have evaluated the generalization performance of our method. Specifically, following RoomFormer \cite{yue2023connecting}, we trained our model on the Structured3D \cite{zheng2020structured3d} training set and then evaluated its performance on the SceneCAD \cite{avetisyan2020scenecad} validation set. We compared our model with existing corner-based methods, HEAT \cite{chen2022heat} and RoomFormer \cite{yue2023connecting}. As shown in Table \ref{tab:generalize}, the performance of the corner-based methods significantly decreases, indicating that relying on local corner features is more sensitive to the changes in data distribution. In contrast, our method learns the global shape features of rooms from the entire image space and is, therefore, more robust to data variations.

\subsection{Ablation Study}
\subsubsection{Pre-processing.}
We evaluate the impact of pre-processing, as shown in the \(2^{\text{nd}}\) row of Table \ref{tab:ablation_1}. After considering the height information of walls, the overall metrics have improved, which indicates that the height information of the wall can enhance the model's perception of wall signals.

\begin{table}[!t]
\centering
\caption{Ablation studies.}
\setlength{\tabcolsep}{4pt}
\begin{tabular}{c cc cc cc}
\hline
\multirow{2}{*}{Settings} & \multicolumn{2}{c}{Room} & \multicolumn{2}{c}{Corner} & \multicolumn{2}{c}{Angle} \\ 
            \cmidrule(r){2-3} \cmidrule(r){4-5} \cmidrule(r){6-7}
    & Prec. & Rec.  & Prec.    & Rec.     & Prec.   & Rec.    \\ \hline
    Default & \textbf{99.5}  & \textbf{98.7}  & \textbf{90.8}  & \textbf{84.9}  & \textbf{89.6}  & \textbf{84.3}  \\
    w/o pre-processing & 99.1 & 97.9 & 88.1 & 84.8 & 89.1 & 83.9 \\
    w/o training strategy & 97.7 & 96.1 & 76.1 & 80.8 & 68.2 & 72.5 
\\ \hline
\end{tabular}
\label{tab:ablation_1}
\end{table}
\begin{table}[!t]
\centering
\caption{Analysis on number of lines and primitives.}
\setlength{\tabcolsep}{4pt}
\begin{tabular}{cc cc cc cc cc}
\hline
\multicolumn{2}{c}{Settings}  & \multicolumn{2}{c}{Room} & \multicolumn{2}{c}{Corner} & \multicolumn{2}{c}{Angle} \\ 
       \cmidrule(r){1-2} \cmidrule(r){3-4} \cmidrule(r){5-6} \cmidrule(r){7-8}
    $l$ & $u$  & Prec. & Rec. & Prec. & Rec. & Prec. & Rec.    \\ \hline
    128 & 64  & 98.1 & 97.2 & 89.4 & 83.0 & 88.2 & 82.3 \\    
    256 & 64  & 99.5 & 98.7 & 90.8 & 84.9 & 89.6 & 84.3  \\ 
    512 & 64  & 99.3 & 98.4 & 90.7 & 85.2 & 89.8 & 84.6 \\ 
    1024 & 64 & 99.1 & 98.1 & 86.9 & 84.5 & 86.8 & 83.6 \\ \hline
    256 & 32  & 98.6 & 97.8 & 89.5 & 84.1 & 88.4 & 83.1 \\
    256 & 64  & 99.5 & 98.7 & 90.8 & 84.9 & 89.6 & 84.3  \\
    256 & 128 & 99.6 & 98.5 & 91.1 & 85.1 & 89.7 & 84.7
\\ \hline
\end{tabular}
\label{tab:ablation_2}
\end{table}
\subsubsection{Training strategy.}
To facilitate the learning of room geometries, we propose to first optimize horizontal and vertical lines and then jointly optimize all lines. As a contrast, we tested the performance of optimizing all lines together. As shown in the \(3^{\text{rd}}\) row of Table \ref{tab:ablation_1}, the model's performance significantly decreases if all lines are optimized simultaneously. Our key insight is that
most rooms can be constructed using axis-aligned lines, which simplifies the problem since these lines require fewer parameters than diagonal lines. This simplification allows the network to learn the basic room shape during the initial training phase. After obtaining the basic shape, diagonal lines are introduced to represent the room's angled areas, enabling a more generalized room shape. We also provide the experiment with only diagonal lines and find it to be less effective, please see the supplementary for more details.




\subsubsection{Line/primitive count.}

We evaluate the effect of different numbers of lines and primitives in Table \ref{tab:ablation_2}. 
As demonstrated in the first third rows, setting lower line parameters leads to a slight decline in performance, while increasing these parameters leads to only marginal improvements. When it comes to 1024 lines, there is a significant drop in performance due to the expanded solution space, making it harder for the model to converge on the optimal solution.
Likewise, as depicted in the last three rows, lower primitive parameters slightly affect performance, whereas higher settings offer negligible improvements. Consequently, following the discussion above, we have set \(l\) to 256 and \(u\) to 64.


\section{Conclusion}
In this work, we present a novel method for floorplan reconstruction. We develop a learning framework called FRI-Net to learn the implicit representation of the floorplan rooms. Our method is capable of generating more plausible room shapes by incorporating geometric priors of the floorplan into our training strategy.

\begin{figure*}[!t]
    \centering
    \includegraphics[width= 0.98\textwidth]{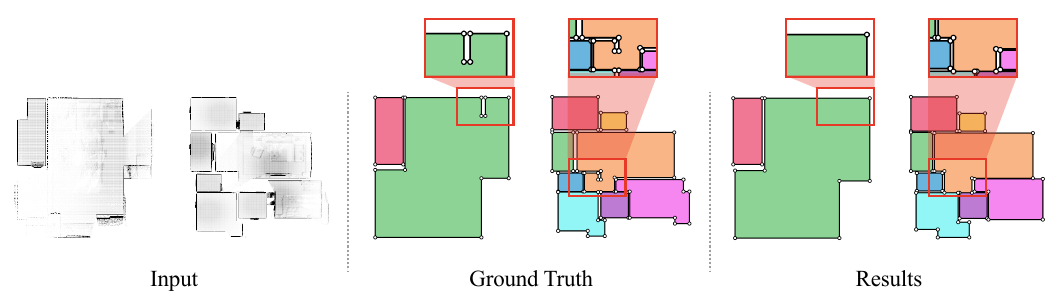}
    \caption{Failure cases. }
    \label{fig:failure cases}
\end{figure*}

However, our method exhibits certain limitations in accurately reconstructing inner walls within rooms. As illustrated in Figure \ref{fig:failure cases}, our method fails to accurately reconstruct the structure of the inner walls. The use of implicit neural representations to learn room shapes, with a focus on global shape features, might result in the over-smoothing of fine local structures. In future work, we aim to combine explicit and implicit representations to precisely capture high-frequency local details while also preserving the overall global structure. 
Furthermore, while the height map extracted from the pre-processing module contains valuable structural cues, its direct application for 3D reconstruction is fraught with challenges due to noise and incomplete data. We will explore its use for 3D floorplan reconstruction in future work.

\section*{Acknowledgements}
We thank the anonymous reviewers for their valuable comments. This work was supported in parts by NSFC (62322207, 62072316), Guangdong Natural Science Foundation (2021B1515020085, 2023A1515011297), Shenzhen Science and Technology Program (RCYX20210609103121030). Guangdong Laboratory of Artificial Intelligence and Digital Economy (SZ) and Scientific Development Funds of Shenzhen University.

%
%
\bibliographystyle{splncs04}
\bibliography{reference}
\end{document}